\documentclass[11pt,a4paper]{article}
\usepackage[hyperref]{emnlp2020}
\usepackage{times}
\usepackage{latexsym}
\usepackage{multirow}
\usepackage{multicol}
\usepackage{graphicx}
\usepackage{bm}

\usepackage{microtype}

\aclfinalcopy 

\usepackage{amsmath}
\usepackage{xcolor}
\usepackage{graphicx}
\usepackage{amsfonts}  
\usepackage{amssymb}
\usepackage{url}
\usepackage{appendix}
\usepackage{cases}
\usepackage{array}
\usepackage{times}
\usepackage{latexsym}
\usepackage{graphicx}
\usepackage{hhline}
\usepackage{cleveref}
\usepackage{arydshln}

\title{The World is Not Binary:\\ Learning to Rank with Grayscale Data for Dialogue Response Selection}
\author{Zibo Lin$^{1,2,}$\footnotemark[1], Deng Cai$^{3,}$\thanks{~~Equal contribution. Work was done during internship at Tencent AI Lab.}~, Yan Wang$^{4}$, Xiaojiang Liu$^4$, Hai-Tao Zheng$^{1,2}$, Shuming Shi$^4$ \\ $^1$Tsinghua Shenzhen International Graduate School, Tsinghua University \\$^2$Department of Computer Science and Technology, Tsinghua University \\$^3$The Chinese University of Hong Kong \\$^4$Tencent AI Lab \\ {\tt \{lzb18,zheng.haitao\}@mails.tsinghua.edu.cn}\\ {\tt thisisjcykcd@gmail.com}{\tt ,} {\tt xiaojiangliu84@hotmail.com}\\ {\tt \{brandenwang,shumingshi\}@tencent.com}}

\date{}

\begin{document}
	\maketitle
	\begin{abstract}
		Response selection plays a vital role in building retrieval-based conversation systems. Despite that response selection is naturally a learning-to-rank problem, most prior works take a point-wise view and train binary classifiers for this task: each response candidate is labeled either relevant (one) or irrelevant (zero). On the one hand, this formalization can be sub-optimal due to its ignorance of the diversity of response quality. On the other hand, annotating grayscale data for learning-to-rank can be prohibitively expensive and challenging. In this work, we show that grayscale data can be automatically constructed without human effort. Our method employs off-the-shelf response retrieval models and response generation models as automatic grayscale data generators. With the constructed grayscale data, we propose multi-level ranking objectives for training, which can (1) teach a matching model to capture more fine-grained context-response relevance difference and (2) reduce the train-test discrepancy in terms of distractor strength. Our method is simple, effective, and universal. Experiments on three benchmark datasets and four state-of-the-art matching models show that the proposed approach brings significant and consistent performance improvements.
	\end{abstract}
	\section{Introduction}
	Building intelligent conversation systems \cite{shum2018eliza, kollar2018alexa} is gaining more and more attention in recent years. A core module in such kind of conversation systems is response selection \cite{ritter2011data,hu2014convolutional,wu2017sequential,tao2019one}: Identifying the best response from a set of possible candidates given a dialogue context, i.e., conversation history. For the response selection problem, the trendy practice is to build neural matching models \cite{ji2014information,wang2015syntax,xu2016incorporating,wu2017sequential,zhou2018multi,lu2019constructing} for scoring the adequacy of individual response candidates in the dialogue context. Most prior works on this topic focus on fine-grained
	text encoding and better interactions between dialogue context and response candidates, typically via sophisticated and powerful matching networks \cite{wu2017sequential,zhou2018multi,lu2019constructing,gu2019interactive}. Despite their differences, in almost all these previous works, the matching models are trained with binary classification objective. Each response in the training data is either labeled positive (i.e., a correct response to the dialogue context) or negative (i.e., an incorrect response). Often, the negative responses are automatically constructed by random sampling.
	\begin{table}
		\begin{center}
			\resizebox{1.0\columnwidth}{!}{
				\begin{tabular}{l|l}
					\hline
					\textbf{Dialogue Context Between Speakers A and B} & \textbf{Relevance} \\
					\hline
					A: Would you please share some useful experience for\\ \hspace{1em} improving spoken English? & \\
					B: Sure! Watching English movies helped a lot. & \\
					A: Agreed. I watched Friends many times. & \\
					B: Me too! I bought the DVDs and they went broken \\ \hspace{1em} due to my frequent use. & \\
					\hline
					\textbf{Ground Truth} & \\
					G: Hah! Then your English should be very good! & $+$\\
					\hline
					\textbf{Distractor Response During Training} & \\
					R1: Why didn't the British police come? & $-$$-$$-$\\
					\hline
					\textbf{Distractor Responses in Real-world Scenario} & \\
					R2: It's said that a DVD can be preserved for decades. & $-$\\
					R3: Friends is an American television sitcom. & $-$\\
					\hline
				\end{tabular}
			}
		\end{center}
		\caption{Dialogue context (conversation history) between Speakers A and B. R1 is a random sample used as a negative instance during training. R2 and R3 are real distractors during testing.}
		\label{retrieved_example}
	\end{table}
	
	One limitation of the above training strategy is that this formalization downplays the nuance of fine-grained response quality; the matching model is only informed to predict a binary label, either correct or incorrect. However, the quality of possible response candidates may be quite diverse, thus letting the matching model be aware of which response candidates are more incorrect or less incorrect than others may more effectively increase the model capacity. Another limitation is that in real-world scenarios the matching models are often confronted with more difficult tasks: to select the best response from a set of strong response candidates instead of random ones. An example is given in Table \ref{retrieved_example}. During training, the matching models are trained to distinguish the ground truth G and the randomly sampled response R1, where R1 shows little relevance to the dialogue context. Matching models trained on such training data have little experience to identify the ground-truth response G from a set of strong distractor responses such as R2 and R3. Intuitively, a good matching model should be able to not only distinguish good responses from random ones (usually totally irrelevant), as conveyed by the binary classification objective, but also capture the more subtle differences for competitive candidates.
	
	One natural solution to the above problems is to collect grayscale data for training; if we consider the quality of all possible response candidates falls in the interval $[0, 1]$, the golden-truth and random responses usually cover the two endpoints only, and our goal is to obtain a list of grayscale responses locate in between $0$ and $1$. However, grayscale data are hard to obtain in reality owing to the expense of human annotation and the subjectivity of individual human annotators.
	
	In this work, we propose to automatically construct grayscale data from standard dialogue datasets, where only golden dialogue context and response pairs are provided. To meet this goal, we resort to off-the-shelf retrieval algorithms and generation models. Our idea is inspired by the observation that, in most cases, the responses from retrieval models or generation models are better than randomly sampled ones but worse than the ground-truth response. We believe that this progressive relationship, such as ``\textit{ground truth $>$ retrieval $>$ random}'', can be utilized for training a better matching model. Concretely, we propose a multi-level ranking objective to make full use of such relationships. Our multi-level ranking objective jointly combines multiple binary contrastive estimations. In addition, the grayscale data partly simulates the real-world response distractors and thus reduces the gap between training and testing, leading to a better distinguishing ability for strong response distractors. 
	
	Our method is simple, effective, and orthogonal to prior efforts for modeling designs. It can be conveniently implemented with most existing matching models. Experimental results on four state-of-the-art matching models and three benchmark datasets demonstrate that our new training approach leads to remarkable performance improvement consistently.
	\section{Background}
	\label{sec:Problem}
	Early research for response selection is devoted to single-turn conversations \cite{wang2013dataset,tan2015lstm,yan2016learning}. Recently, researchers have started to study on multi-turn conversations \cite{lowe2015ubuntu,wu2017sequential,zhang2018modeling}. In the current literature, the task of response selection is formulated as follows. Given a dialogue dataset $\mathcal{D} = \{(c_i, r_i)\}$, where $c_i$ represents a dialogue context, and $r_i$ is the human-written ground-truth response. The goal is to build a matching model $s(\cdot,\cdot)$ from $\mathcal{D}$ so that $s(c,r)$ accurately measures the adequacy of a response candidate $r$ for a dialogue context $c$.
	\begin{figure*}[t]
		\centering
		\includegraphics[width=0.8\linewidth]{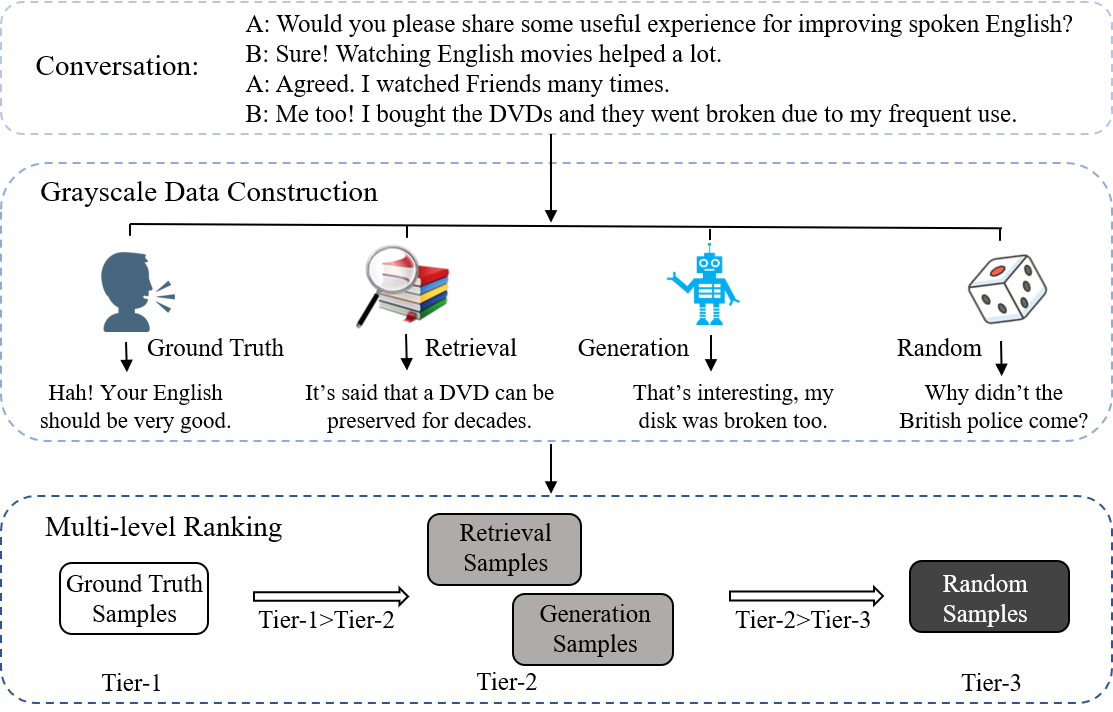}
		\caption{The illustration of our training approach. For each dialogue, we first extract a number of grayscale data from heterogeneous sources. Then, the multi-level ranking objective is applied to learn the progressive relationships between different responses.}
		\label{framework_pic}
	\end{figure*}
	
	Rapid progress has been made for building such matching models in recent years. Concretely, various neural architectures \cite{zhou2016multi,wu2017sequential,zhou2018multi,gu2019interactive,tao2019one,yuan2019multi} have been proposed for fine-grained
	text encoding and better dialogue context and response interactions modeling. To train such matching models, binary-labeled training sets are constructed \cite{lowe2015ubuntu,wu2017sequential,zhang2018modeling}: The human-written ground-truth response is designated as positive instances (labeled as 1), and a set of randomly sampled responses $\mathcal{N}_i$ are treated as negative ones (labeled as 0). The learning objective of $s(\cdot,\cdot)$ is then to maximize the following binary classification loss function:
	\begin{equation}
	\label{cross_entroy}
	\log{s(c_i,r_i)} + \mathbb{E}_{r^-\in\mathcal{N}_i}\log{(1-s(c_i,r^-))}.
	\end{equation}
	
	Different from previous works, our study questions the effectiveness of the binary-labeled training data and the corresponding binary classification objective. We argued that the binary classification paradigm is sub-optimal as most of the randomly sampled negative responses are distant from the corresponding positive responses in terms of matching degree, which could lead to serious drawbacks when some strong distractors are presented during testing \cite{zhou2018multi,zhang2018modeling}. Our work starts with enriching the range of the negative sample set $\mathcal{N}_i$ in terms of response quality and leads to a simple but new learning strategy that aimed at capturing more fine-grained response quality differences.
	\section{Proposed Approach}
	\subsection{Overview}
	Figure \ref{framework_pic} depicts an overview of our approach. First, different responses are acquired from various sources, such as retrieval models, generation models, and random sampling. Then, the collected responses are sorted by estimated quality to form progressive relationships. Lastly, a multi-level ranking objective is designed to learn such relationships. We first present our methods for automatically constructing grayscale data in Section \ref{sec:Construction}, followed by the multi-level ranking objective introduced in Section \ref{sec:objective}.
	\subsection{Grayscale Data Acquisition}
	\label{sec:Construction}
	Our goal is to construct a set of responses with diverse quality. Specifically, we construct three types of responses for each dialogue context and rank them in three tiers. It should be noted that our data acquisition only relies on standard dialogue datasets, which only provide human-to-human dialogue context and response pairs. 
	\paragraph{Zero \& One} First of all, the corresponding responses for dialogues context in the standard dialogue dataset are considered as our ground-truth responses. These human-written responses are often informative and relevant. As a result, the ground-truth samples are ranked as tier-1. Similar to previous work, we also utilize randomly sampled responses for contrastive estimation. The random responses are sampled from the responses of other dialogue contexts in the training data. We rank random responses as tier-3 because they often show little relevance to the dialogue context. The ground-truth responses and random responses constitute the ``zero \& one'' binary training data used in the prior work.
	
	\paragraph{Grayscale} We now delve into describing the grayscale data construction procedures. We consider two types of frequently-used toolkits for automatic response generation to produce grayscale data, namely, the retrieval-based models and the generation-based models.
	
	The retrieval-based models \cite{ji2014information,hu2014convolutional}  directly copy an existing response from the training corpus when receiving a response request. Since the returning responses are always human utterances in real-world conversations, they are informative and grammatical. However, the response quality of such systems varies as it depends on the lexical similarity of the given dialogue context and those in the training corpus. Typically, the retrieval results are better than random responses because they are more or less relevant to the dialogue context. However, most retrieval results are worse than the ground truth. The retrieval results are ranked tier-2.
	
	Specifically, we split the multi-turn dialogue into a series of single-turn input-response pairs. Then we index the input-response pairs with the BM25 algorithm \cite{robertson2009probabilistic}. We retrieve response candidates using the last utterance of the dialogue context.
	
	The generation-based models \cite{shang2015neural,li-etal-2016-diversity} generate a new utterance from scratch after training. While those models have better generalization capacity in rare dialogue contexts, the generation responses tend to be universal and noninformative (e.g., “I don’t know”, “I think so” etc.) \cite{li-etal-2016-diversity}. Similar to the retrieval responses, the generation responses are usually better than the random responses but worse than the ground-truth responses. However, compared to retrieval models that merely rely on lexical overlapping, generation results can capture deeper semantic interactions. The different characteristics of retrieval and generation models make their results complement each other in terms of response quality, which we consider beneficial for training.
	
	Specifically, we train a Seq2Seq model with the attention mechanism \cite{bahdanau2015Neural} for response generation. We adopt the same corpus used in the retrieval model to train the generation model. The generation response is produced by feeding the dialogue context to a trained model.
	\paragraph{Discussion on Extendibility}
	Note that there can be many more sophisticated ways to construct the grayscale data. For example, one may employ the results from different retrieval models and/or generation models. Responses from different models can be further divided into sub-groups according to the relative strengths of the corresponding models. For instance, responses that are generated from more advanced and competent generation models (e.g., a model based on GPT2 \cite{radford2019language}) can be considered better than those from less competent models (e.g., a vanilla seq2seq model). However, in this paper,  we only showcase the results with basic retrieval and generation models for keeping our idea simple and neat. Nevertheless, this simple setting, as we will demonstrate, already leads to remarkable performance improvements.
	\subsection{Multi-Level Ranking Objectives}
	\label{sec:objective}
	Our grayscale data acquisition provides ground for carrying out more principled and sufficient training paradigms. To make full use of the grayscale data, we propose multi-level ranking objectives. Unlike prior work that minimizes binary classification errors, our training objective better fits the learning-to-rank nature of response selection, that is, minimizes ranking errors of possible responses \cite{cao2007learning}. Also, as the grayscale data exhibit various response quality, training with such data rather than random negatives better simulate testing environments.
	
	We start formal descriptions with some notation: the training set can be re-organized as $\overline{\mathcal{D}} = \{(c_i, R_i)\}_{i=1}^N$, where $c_i$ denotes the dialogue context and $R_i=\{r_i, e_i, g_i, \tilde{r}_i\}$ is the response set enhanced by grayscale data. Concretely, $r_i$, $e_i$, $g_i$, and $\tilde{r}_i$ refer to ground-truth responses, retrieval responses, generation responses, and random responses, respectively. We consider three ordered list as follows.
	\begin{itemize}
		\item{\textit{ground truth$ > $retrieval$ > $random}} This ordered list considers the progressive relationships between ground-truth responses, retrieval responses, and random responses. We use margin ranking losses for implementation, the formula are given below:
		\begin{equation} \nonumber
		\label{retrievalpath}
		\begin{split}
		\mathcal{L}_{Ret} =&\max\{0, \mu - s(c, r_i) + s(c, e_i)\}\\
		+& \max\{0, \mu - s(c,e_i) + s(c, \tilde{r}_i)\}.
		\end{split}
		\end{equation}
		where $\mu$ is a hyperparameter and represent the minimum acceptable score margin between two tiers, and $s(\cdot,\cdot)$ is the matching score given by a matching model.
		
		\item{\textit{ground truth$>$generation$>$random}} This ordered list considers the progressive relationships between ground-truth responses, generation responses, and random responses. The loss function is given below.
		\begin{equation} \nonumber
		\label{generationpath}
		\begin{split}
		\mathcal{L}_{Gen} =&\max\{0, \mu - s(c, r_i) + s(c, g_i)\}\\ 
		+ &\max\{0, \mu - s(c,g_i) + s(c, \tilde{r}_i)\},
		\end{split}
		\end{equation}
		
		\item{\textit{ground truth$>$random}}
		\begin{equation} \nonumber
		\label{randompath}
		\mathcal{L}_{Ran} =\max\{0, \mu - s(c, r_i) + s(c, \tilde{r}_i)\},
		\end{equation}
		this loss function directly models the relationship between the ground-truth samples $r_i$ and random samples $\tilde{r}_i$.
	\end{itemize}
	
	Our final training objective is an unite of all above. It models the integrated relationship between tiers ``\textit{ground truth$>$retrieval $\&$ generation$>$random}'' and ``\textit{ground truth $>$ random}'' simultaneously:
	\begin{equation} \nonumber
	\label{multipath}
	\mathcal{L}_{Uni} = \mathcal{L}_{Ran} + \mathcal{L}_{Ret} + \mathcal{L}_{Gen}.
	\end{equation}
	\section{Experimental Setup}
	\subsection{Datasets and Evaluation Metrics}
	We test on three benchmark datasets for multi-turn response selection.
	\paragraph{Ubuntu Dialogue Corpus} It consists of English multi-turn dialogues about technical support collected from the Ubuntu Forum \cite{lowe2015ubuntu}. The dataset contains 500K, 50K and 50K chat logs for training, validation, and test respectively. Each test dialogue is paired with 9 distractor responses. Following conventions, the response selection performance is evaluated by $R_n$@$k$ scores. $R_n$@$k$ is the recall rate at position $k$ in $n$ candidates.
	
	\paragraph{Douban Conversation Corpus} It consists of Chinese multi-turn daily conversations crawled from Douban group \cite{wu2017sequential}. The dataset contains 500K, 25K and 1K chat logs for training, validation, and test respectively. Each test dialogue is paired with 10 candidate responses. Following prior work, besides $R_n$@$k$ scores, we also report Mean Average Precision (MAP), Mean Reciprocal Rank (MRR) and the precision at position 1 (P@$1$).
	
	\paragraph{E-commerce} It consists of Chinese conversations between customers and customer service staff from Taobao \cite{zhang2018modeling}. The dataset sizes and settings is the same as Douban corpus. $R_n$@$k$ scores are commonly employed for evaluation.
	\subsection{Baseline Models}
	We compare with the following baseline models.
	\begin{table*}[t]
		\centering
		\resizebox{2.0\columnwidth}{!}{
			\begin{tabular}{l||cccccc||cccc||ccc}
				\multicolumn{1}{l||}{\multirow{2}{*}{\textbf{Model}}}  & \multicolumn{6}{c||}{\textbf{Douban}} & \multicolumn{4}{c||}{\textbf{Ubuntu}}  & \multicolumn{3}{c}{\textbf{E-commerce}} \\ \cline{2-14} 
				& ${MAP}$ & $ MRR$ & $ P$@$1$ & $ R_{10}$@$1$ & $ R_{10}$@$2$ & $ R_{10}$@$5$  & $ R_2$@$1$ & $ R_{10}$@$1$ & $ R_{10}$@$2$ & $ R_{10}$@$5$  & $ R_{10}$@$1$ & $ R_{10}$@$2$ & $ R_{10}$@$5$ \\
				\hline
				\hline
				RNN & 0.390 & 0.422 & 0.208 & 0.118 & 0.223 & 0.589   & 0.768 & 0.403 & 0.547 & 0.819   & 0.325 & 0.463 & 0.775 \\
				CNN & 0.417 & 0.440 & 0.226 & 0.121 & 0.252 & 0.647   & 0.848 & 0.549 & 0.684 & 0.896   & 0.328 & 0.515 & 0.792 \\
				LSTM & 0.485 & 0.527 & 0.320 & 0.187 & 0.343 & 0.720   & 0.901 & 0.638 & 0.784 & 0.949   & 0.365 & 0.536 & 0.828 \\
				BiLSTM & 0.479 & 0.514 & 0.313 & 0.184 & 0.330 & 0.716   & 0.895 & 0.630 & 0.780 & 0.944   & 0.355 & 0.525 & 0.825 \\
				MV-LSTM & 0.498 & 0.538 & 0.348 & 0.202 & 0.351 & 0.710    & 0.906 & 0.653 & 0.804 & 0.946    & 0.412 & 0.591 & 0.857 \\
				Match-LSTM & 0.500 & 0.537 & 0.345 & 0.202 & 0.348 & 0.720    & 0.904 & 0.653 & 0.799 & 0.944    & 0.410 & 0.590 & 0.858 \\
				\hline
				DL2R & 0.488 & 0.527 & 0.330 & 0.193 & 0.342 & 0.705  & 0.899 & 0.626 & 0.783 & 0.944  & 0.399 & 0.571 & 0.842 \\
				Multi-View & 0.505 & 0.543 & 0.342 & 0.202 & 0.350 & 0.729  & 0.908 & 0.662 & 0.801 & 0.951  & 0.421 & 0.601 & 0.861 \\
				DUA & 0.551 & 0.599 & 0.421 & 0.243 & 0.421 & 0.780   & - & 0.752 & 0.868 & 0.962  & 0.501 & 0.700 & 0.921 \\
				\hline
				SMN & 0.529 & 0.569 & 0.397 & 0.233 & 0.396 & 0.724   & 0.926 & 0.726 & 0.847 & 0.961  & 0.453 & 0.654 & 0.886 \\
				DAM & 0.550 & 0.601 & 0.427 & 0.254 & 0.410 & 0.757   & 0.938 & 0.767 & 0.874 & 0.969  & 0.526 & 0.727 & 0.933 \\ 
				IOI & 0.573 & 0.621 & 0.444 & 0.269 & 0.451 & 0.786   & 0.947 & 0.796 & 0.894 & 0.974  & 0.563 & 0.768 & 0.950 \\
				MSN & 0.587 & 0.632 & 0.470 & 0.295 & 0.452 & 0.788   & - & 0.800 & 0.899 & 0.978  & 0.606 & 0.770 & 0.937 \\
				\hline
				G-SMN & 0.564 & 0.615 & 0.443 & 0.271 & 0.439 & 0.781   & 0.938 & 0.765 & 0.873 & 0.969   & 0.504 & 0.713 & 0.926 \\
				G-DAM & 0.588 & 0.637 & 0.464 & 0.284 & 0.466 & 0.822   & 0.946 & 0.789 & 0.891 & 0.986   & 0.564 & 0.769 & 0.948 \\
				G-IOI & 0.591 & 0.639 & 0.454 & 0.277 & 0.458 & 0.796   & 0.951 & 0.805 & 0.902 & 0.981   & 0.579 & 0.772 & 0.955 \\
				G-MSN & \textbf{0.599} & \textbf{0.645} & \textbf{0.476} & \textbf{0.308} & \textbf{0.468} & \textbf{0.826}   & \textbf{0.958} & \textbf{0.812} & \textbf{0.911} & \textbf{0.987}   & \textbf{0.613} & \textbf{0.786} & \textbf{0.964} \\
			\end{tabular}
		}
		\caption{\label{dam_result_table} Evaluation results of all models trained with our approach on Douban, Ubuntu and, E-commerce datasets. Results of all baselines are directly copied from the previous works \cite{tao2019one,yuan2019multi}.}
	\end{table*}
	\begin{table*}[t]
		\small
		\begin{center}
			\begin{tabular}{c|c|c||cccc||cccc} 
				\multicolumn{1}{c|}{\multirow{2}{*}{$\mathcal{L}_{Ran}$}} &
				\multicolumn{1}{c|}{\multirow{2}{*}{$\mathcal{L}_{Ret}$}} &  \multicolumn{1}{c||}{\multirow{2}{*}{$\mathcal{L}_{Gen}$}}  & \multicolumn{4}{c||}{\textbf{SMN}} & \multicolumn{4}{c}{\textbf{DAM}}\\ 
				\cline{4-11}
				&& & $ P$@$1$ & $ R_{10}$@$1$ & $ R_{10}$@$2$ & $ R_{10}$@$5$   & $ P$@$1$ & $ R_{10}$@$1$ & $ R_{10}$@$2$ & $ R_{10}$@$5$  \\
				\hline
				\hline
				\checkmark &$\times$ & $\times$ & 0.403 & 0.240 & 0.418 & 0.768   & 0.423 & 0.253 & 0.435 & 0.784   \\
				\hdashline
				\checkmark &$\times$ & \checkmark & 0.421 & 0.256 & 0.410 & 0.772    & 0.439 & 0.266 & 0.449 & 0.788    \\
				\checkmark &\checkmark & $\times$ & 0.439 & 0.267 & 0.431 & 0.768    & 0.449 & 0.270 & 0.447 & 0.801    \\
				\hdashline
				\checkmark &\checkmark & \checkmark & \textbf{0.443} & \textbf{0.271} & \textbf{0.439} & \textbf{0.781}    & \textbf{0.464} & \textbf{0.284} & \textbf{0.466} & \textbf{0.822}   \\
			\end{tabular}
		\end{center}
		\caption{\label{ablation_table} Ablation study of our approach on Douban datasets with SMN and DAM.}
	\end{table*}
	\paragraph{Single-turn Matching Models} These models concatenate all context utterances together into one single long utterance then compute the matching scores between the long utterance and response candidates, including RNN~\cite{lowe2015ubuntu}, CNN~\cite{lowe2015ubuntu}, LSTM~\cite{lowe2015ubuntu}, Bi-LSTM~\cite{kadlec2015improved}, Match-LSTM~\cite{wang2016learning} and MV-LSTM~\cite{wan2016match}.
	\paragraph{Multi-turn Matching Models} These models aggregate the information of context utterances in more advanced ways, including DL2R~\cite{yan2016learning}, Multi-View~\cite{zhou2016multi}, DUA~\cite{zhang2018modeling}, SMN~\cite{wu2017sequential}, DAM~\cite{zhou2018multi}, IOI~\cite{tao2019one}, and MSN~\cite{yuan2019multi}.
	\subsection{Implementation Details}
	For grayscale data construction, we train a seq2seq generation model and build a BM25 retrieval system using the training set for each dataset. We consider the top 100 responses from BM25 retrieval and the top 5 responses from seq2seq generation (via beam search) as the grayscale responses. To facilitate further research, we have made our collected grayscale data publicly available.\footnote{Related resources can be found at \url{https://ai.tencent.com/ailab/nlp/dialogue/datasets/grayscale_data_release.zip}} During training, we use these grayscale responses in a way adaptive to the training matching model. At each training epoch, ten different grayscale responses are used: the top 5 retrieval responses ranked by the current matching model and all 5 seq2seq generation responses. We experiment our new training approach on four latest state-of-the-art models as follows:
	\begin{itemize}
		\item{SMN}~\cite{wu2017sequential} interacts each utterance of a dialogue context with a response and then transforms interaction matrices into matching vectors with CNN. The matching vectors are finally mapped into a matching score with an RNN.
		\item{DAM}~\cite{zhou2018multi} obtains matching vectors of text segments at different granularities with the stacked self-attention. The matching vectors are then distilled with the cross-attention and finally fused into a matching score via a single-layer perceptron.
		\item{IOI}~\cite{tao2019one} pairs each utterance of a context with a response via stacking multiple interaction blocks and then aggregates matching information from all the pairs as a matching score in an iterative fashion.
		\item{MSN}~\cite{yuan2019multi} utilizes a multi-hop selector to select the relevant utterances as context and then matches the filtered context with the given response candidate to obtain a matching score.
	\end{itemize}
	Specifically, we first pre-train a model with objective $\mathcal{L}_{ran}$ only then switch to $\mathcal{L}_{Uni}$. We find that such a treatment makes the training process more stable.
	
	\section{Results and Discussion}
	\subsection{Experimental Results}
	The experimental results are listed in Table \ref{dam_result_table}, where G-X indicates X with our grayscale enhanced training approach. We can see that our training approach significantly improves the performance of all four matching models in terms of various metrics. The improvements are consistent across different datasets and different models, indicating the university of our approach. Moreover, one interesting observation is that a less-accurate matching architecture with the proposed training approach can outperform a stronger matching architecture with the traditional training paradigm, e.g., G-IOI vs. MSN. This suggests that while the choice of learning objective is often overlooked, it could be decisive for building a competitive response selection model.
	\subsection{Effect of Different Grayscale Data}
	We then turn to conduct an ablation study for understanding the roles of different grayscale data in performance enhancement. We choose SMN as well as DAM as the baselines models. We train the models with three additional settings by removing either retrieval responses or generation responses and removing both of them. 
	
	The results are shown in Table \ref{ablation_table}, we can find that both retrieval data and generation data make irreplaceable contributions to the overall performance and the combination of both worlds makes the best results, which confirms our hypotheses that responses from heterogeneous sources complement each other. We can also find that the help from retrieval data has a greater influence than generation data when used alone. This can be attributed to that the seq2seq-based generation model tends to output general and dull responses. Such general responses are less informative than the retrieval data, thus can provide limited
	help for distinguishing the nuance of fine-grained response quality. 
	\subsection{Effect of Multi-level Ranking Objectives}
	Next, we study the effect of the multi-level ranking objective (MRO). Recall that we adopt the MRO in order to make use of the progressive relationship in different tiers. However, a simpler alternative is to treat all grayscale data as negative samples and use the learning objective in Eq. \ref{randompath}. It can be regarded as a simple data augmentation technique, enlarging the set of negative examples with retrieval and generation results. We implement such an idea to test whether the proposed MRO is necessary and quantify the benefit of the MRO.
	
	As shown in Table \ref{multi_leval_table}, the performance of models trained without MRO falls behind those trained with MRO. Besides, the improvements of grayscale data without MRO are quite limited compared to the original counterparts without grayscale data. This indicates that the proposed multi-level ranking objective is essential for performance improvement.
	\begin{table}[t]
		\small
		\begin{center}
			\begin{tabular}{c||cccc} 
				\textbf{Model} & $P$@$1$ & $R_{10}$@$1$ & $R_{10}$@$2$ & $R_{10}$@$5$ \\
				\hline
				\hline
				G-SMN  & \textbf{0.443} & \textbf{0.271} & \textbf{0.439} & \textbf{0.781} \\
				$-$\textit{MRO} & 0.410 & 0.244 & 0.416 & 0.766 \\
				\hdashline
				SMN & 0.397 & 0.233 & 0.396 & 0.724 \\
				\hline
				G-DAM   & \textbf{0.464} & \textbf{0.284} & \textbf{0.466} & \textbf{0.822} \\
				$-$\textit{MRO} & 0.427 & 0.252 & 0.422 & 0.782 \\
				\hdashline
				DAM & 0.427 & 0.254 & 0.410 & 0.757 \\
				\hline
				G-IOI & \textbf{0.454} & \textbf{0.277} & \textbf{0.458} & \textbf{0.796} \\
				$-$\textit{MRO} & 0.449 & 0.271 & 0.449 & 0.788 \\
				\hdashline
				IOI & 0.444 & 0.269 & 0.451 & 0.786 \\
				\hline
				G-MSN  & \textbf{0.476} & \textbf{0.308} & \textbf{0.468} & \textbf{0.826} \\
				$-$\textit{MRO} & 0.471 & 0.297 & 0.452 & 0.789 \\
				\hdashline
				MSN  & 0.470 & 0.295 & 0.452 & 0.788 \\
			\end{tabular}
		\end{center}
		\caption{\label{multi_leval_table} Effect of multi-level ranking objectives. Here, all metrics are evaluated in Douban corpus.}
	\end{table}
	\subsection{Effect of Margin Size}
	The hyperparameter margin size ($\mu$) denotes the minimum distance between two tiers in matching scores, which may affect the performance of a matching model. We conduct a series of sensitivity analysis experiments to study how the margin affects the performance of our training.\footnote{We also tried to use different margins for different pairs but the improvements are limited.} All models are evaluated in terms of $R_{10}$@$1$.
	
\begin{figure}[t]
		\centering
		\includegraphics[scale=0.22]{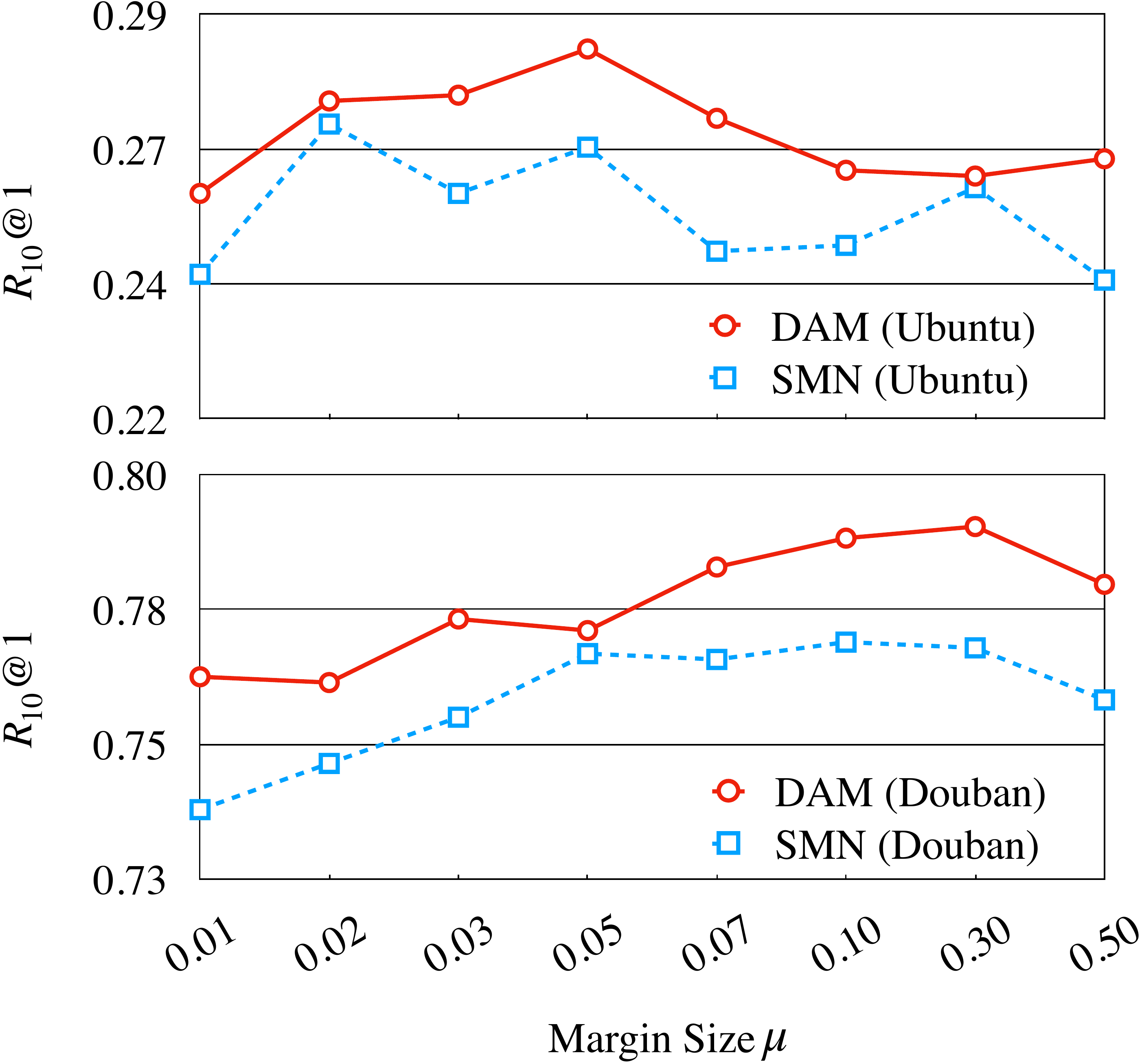}
		\caption{ The effect of margin size.}
		\label{mu-pic}
	\end{figure}

	Referring to Figure \ref{mu-pic}, we can see that both SMN and DAM have a similar trend on Douban: the curves first increase and then drop as the margin increases. This is mainly because response candidates on Douban are of high relevance. When the margin is too large, matching models have no idea to handle strongly relevant distractors. However, when the margin is too small, matching models will become too sensitive and sometimes mistakenly give high scores for responses with less relevance to dialogue context. Results on Ubuntu show a completely different behavior: the performances grow in step with the margin. The reason may be that the response distractors of Ubuntu have relatively large margins in semantic and matching models need to make strong discrimination between the ground truth and other grayscale samples. As a result, models learned with the large margin can fit such data distribution.
	\begin{table*}[t]
		\small
		\begin{center}
			\begin{tabular}{l||cccccc||cccc}
				\multicolumn{1}{l||}{\multirow{2}{*}{\textbf{Model}}}  & \multicolumn{6}{c||}{\textbf{Douban}} & \multicolumn{4}{c}{\textbf{Ubuntu}} \\ \cline{2-11} 
				& $ MAP$ & $ MRR$ & $ P$@$1$ & $ R_{10}$@$1$ & $ R_{10}$@$2$ & $ R_{10}$@$5$  & $ R_2$@$1$ & $ R_{10}$@$1$ & $ R_{10}$@$2$ & $ R_{10}$@$5$ \\ 
				\hline
				SMN & 0.529 & 0.569 & 0.397 & 0.233 & 0.396 & 0.724   & 0.926 & 0.726 & 0.847 & 0.961\\ 
				SMN+CoT & 0.559 & 0.601 & 0.424 & 0.260 & 0.426 & 0.764   & 0.933 & 0.759 & 0.862 & 0.961\\ 
				\hdashline
				G-SMN & 0.564 & 0.615 & 0.443 & 0.271 & 0.439 & 0.781   & 0.938 & 0.765 & 0.873 & 0.969\\
				G-SMN+CoT & \textbf{0.569} & \textbf{0.622} & \textbf{0.458} & \textbf{0.278} & \textbf{0.442} & \textbf{0.793}   & \textbf{0.942} & \textbf{0.771} & \textbf{0.875} & \textbf{0.970} \\
				\hline 
				DAM & 0.550 & 0.601 & 0.427 & 0.254 & 0.410 & 0.757   & 0.938 & 0.767 & 0.874 & 0.969\\ 
				DAM+CoT & 0.583 & 0.628 & 0.451 & 0.276 & 0.454 & 0.806   & 0.944 & 0.782 & 0.884 & 0.967\\ 
				\hdashline
				G-DAM & 0.588 & \textbf{0.637} & \textbf{0.464} & 0.284 & \textbf{0.466} & \textbf{0.822}   & 0.946 & 0.789 & 0.891 & \textbf{0.986}\\
				G-DAM+CoT & \textbf{0.589} & 0.636 & \textbf{0.464} & \textbf{0.286} & 0.464 & 0.821   & \textbf{0.951} & \textbf{0.796} & \textbf{0.892} & 0.981\\
			\end{tabular}
		\end{center}
		\caption{\label{coteach_result_table} Experimental results of matching models trained with our approach and the co-teaching framework. X+CoT indicates models trained with the co-teaching framework. We copy the results of SMN+CoT and DAM+CoT from \citet{feng2019learning} on Douban, and we supplement the results of two models trained with the co-teaching framework on Ubuntu.} 
	\end{table*}
	\begin{table*}[t]
		\centering
		\includegraphics[width=0.94\linewidth]{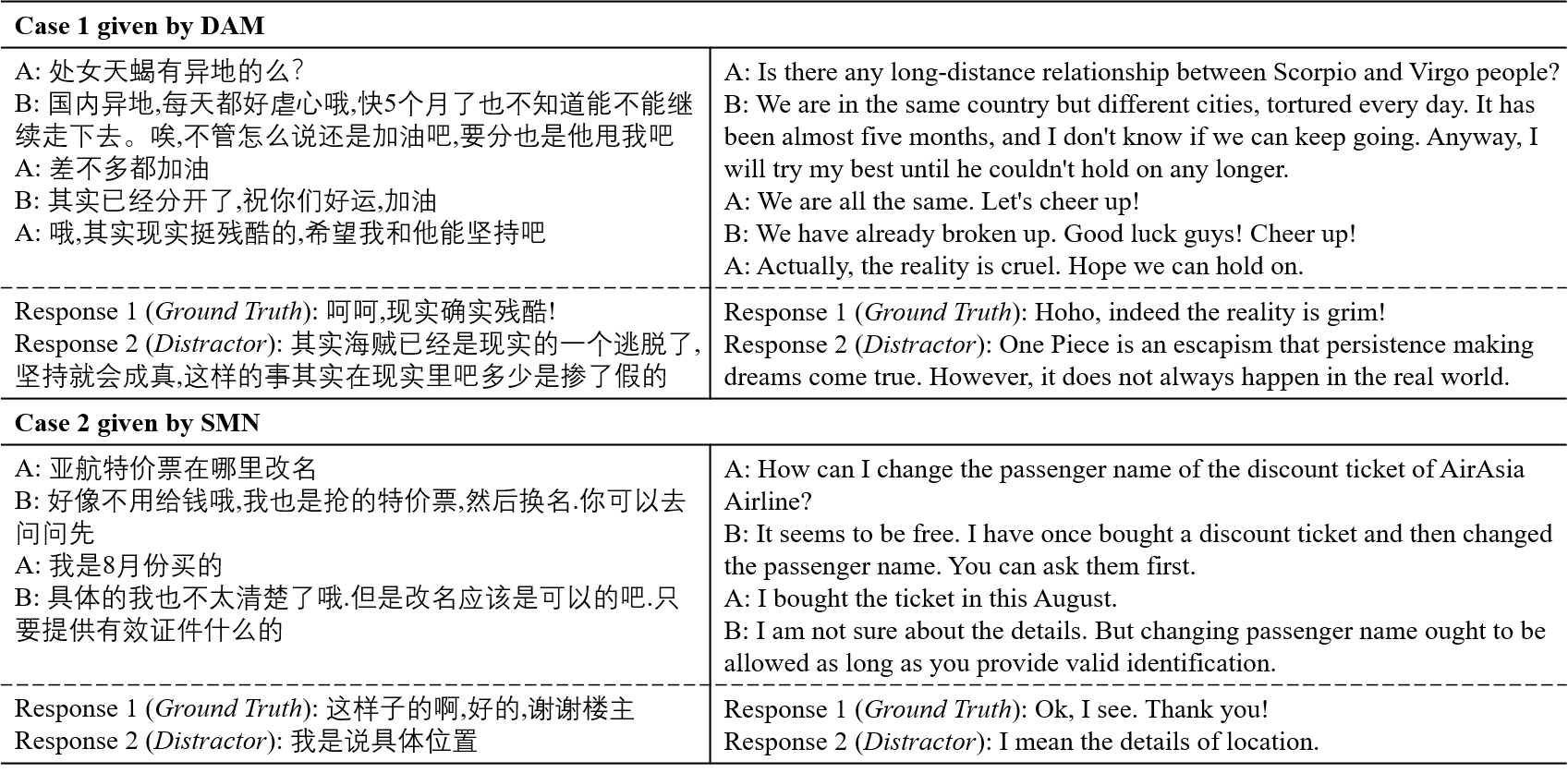}
		\caption{Two cases from the test set of Douban are listed above, and both of them have Response 1 as a ground-truth response. Though each dialogue has ten candidates, we show only two of them due to space limitations. The dialogues are in Chinese (the left) and we also provide their translated version in English (the right).}
		\label{case_study}
	\end{table*}
	
	\subsection{Compatiblity with Co-teaching} We have noticed that \citet{feng2019learning} adopts the co-teaching framework to train a robust matching model. From their experiment, the co-teaching framework with dynamic margins is proven to eliminate the effect from random sampled noisy responses effectively. We believe that our approach and co-teaching framework can benefit each other. Therefore, we combine our training approach with the co-teaching framework taking margins strategy as an instance to train the matching models.
	
	From the results in Table \ref{coteach_result_table}, we can see that models trained with our approach outperform those trained with the co-teaching framework. More importantly, the SMN+CoT and DAM+CoT obtain further improvements after adding our multi-level ranking objectives. This demonstrates that our approach is compatible with the co-teaching framework and shows strong portability and practicability to act as a generalized approach.
	\subsection{Case Study} As shown in case 1 of Table \ref{case_study}, response 2 contains some irrelevant content about the comic ``One Piece'', but it is still selected by DAM as the best response. In case 2, SMN selects the totally irrelevant response 2 as the best response, which may because this response has some overlapped words with the dialogue. These are consistent with the problem introduced in Section \ref{sec:Problem} that these models may mistake the fuzzy-candidate with few improper details for the best response due to the gap between training and testing. In contrast, after adopting our training approach, the G-SMN and G-DAM correctly identify the improper content in the negative responses and successfully select response 1 as the best response.
	\section{Related Work}
	Some researchers also studied how to improve the performance of existing matching models with a better learning method. \citet{wu2018learning} proposed to leverage a Seq2Seq model as a weak annotator to assign a score for each response candidate of the dialogue and learn matching models through the scores. \citet{feng2019learning} introduced the co-teaching framework \cite{han2018co} for eliminating the effect of training noises. The learning approach maintains two matching models and makes them teach each other. \citet{li2019sampling} attempted to neglect the effect of false negatives and trivial true responses by adopting four negative sampling strategies to choose negative samples during training dynamically. Different from those previous works, our approach makes use of grayscale data from heterogeneous sources and learns progressive quality relationships. In addition, our work enhances retrieval models with generation models, which is on par with recent attempts \cite{cai2019skeleton,cai2019retrieval} to strengthen generation models via retrieval models.
	\section{Conclusions}
	We presented a novel approach for training response selection models for multi-turn conversations. It automatically constructs different types of grayscale data and uses a multi-level ranking objective. The proposed approach can teach a matching model to capture fine-grained quality differences better and reduce the train-test discrepancy in distractor strength. Experimental results on three benchmark datasets and four state-of-the-art models demonstrated the effectiveness of the proposed training approach.
	\section*{Acknowledgments}
	This research is supported by National Natural Science Foundation of China (Grant No. 61773229 and 61972219), the Basic Research Fund of Shenzhen City (Grand No. JCYJ20190813165003837), Tencent AI Lab Rhino-Bird Focused Research Program (No. JR202032) and Overseas Cooperation Research Fund of Graduate School at Shenzhen, Tsinghua University (Grant No. HW2018002).
	\bibliography{emnlp2020}
	\bibliographystyle{acl_natbib}
\end{document}